\documentclass[preprint,12pt, a4paper]{elsarticle}

\usepackage{amssymb}
\usepackage{hyperref}
\usepackage{comment}
\usepackage{cite}
\usepackage{amsmath,amsthm,amssymb,amsfonts}
\usepackage{graphicx}
\usepackage{textcomp}
\usepackage{xcolor}
\usepackage[T1]{fontenc}
\usepackage{booktabs}
\usepackage{longtable}
\usepackage{pbox}
\usepackage{mathtools}
\usepackage{caption}
\usepackage{subcaption}
\usepackage{algpseudocode}
\usepackage{algorithm}
\usepackage{siunitx}
\usepackage[inline]{enumitem}
\usepackage{siunitx}
\usepackage{diagbox}
\usepackage[]{mdframed}
\usepackage{listings}
\definecolor{codegreen}{rgb}{0,0.6,0}
\definecolor{codegray}{rgb}{0.5,0.5,0.5}
\definecolor{codepurple}{rgb}{0.58,0,0.82}
\definecolor{backcolour}{rgb}{0.95,0.95,0.92}
\lstdefinestyle{mystyle}{
    backgroundcolor=\color{backcolour},   
    commentstyle=\color{codegreen},
    keywordstyle=\color{magenta},
    numberstyle=\tiny\color{codegray},
    stringstyle=\color{codepurple},
    basicstyle=\ttfamily\footnotesize,
    breakatwhitespace=false,         
    breaklines=true,                 
    captionpos=b,                    
    keepspaces=true,                 
    numbers=left,                    
    numbersep=5pt,                  
    showspaces=false,                
    showstringspaces=false,
    showtabs=false,                  
    tabsize=2
}
\newcolumntype{L}[1]{>{\raggedright\arraybackslash}p{#1}}
\newcolumntype{C}[1]{>{\centering\arraybackslash}p{#1}}
\newcolumntype{R}[1]{>{\raggedleft\arraybackslash}p{#1}}
\theoremstyle{plain}

\usepackage{xparse}
\NewDocumentCommand\N{sm}{\mathcal{N}\IfBooleanT#1{^{\ast}}_{#2}}

\makeatletter
\NewDocumentCommand{\@Coefficients}{m}{\text{\ttfamily\upshape #1}}
\newcommand\uMultilevelCoefficients{\@Coefficients{u\char`_mc}}
\newcommand\newMultilevelCoefficients{\@Coefficients{\~{u}\char`_mc}}
\makeatother

\journal{Software Impacts}

\begin{document}

\begin{frontmatter}



\title{MGARD: A multigrid framework for high-performance, error-controlled data compression and refactoring}


\author[1]{Qian Gong}
\author[2]{Jieyang Chen}
\author[6]{Ben Whitney}
\author[3]{Xin Liang}
\author[1]{Viktor Reshniak}
\author[4]{Tania Banerjee}
\author[4]{Jaemoon Lee}
\author[4]{Anand Rangarajan}
\author[5]{Lipeng Wan}
\author[1]{Nicolas Vidal}
\author[8]{Qing Liu}
\author[1]{Ana Gainaru}
\author[1]{Norbert Podhorszki}
\author[1]{Richard Archibald}
\author[4]{Sanjay Ranka}
\author[1]{Scott Klasky}

\address[1]{Oak Ridge National Laboratory, USA}
\address[2]{University of Alabama at Birmingham, USA}
\address[3]{University of Kentucky, USA}
\address[4]{University of Florida, USA}
\address[5]{Georgia State University, USA}
\address[6]{University of Wisconsin Eau Claire}
\address[8]{New Jersey Institute of Technology, USA}
\begin{abstract}
We describe MGARD, a software providing MultiGrid Adaptive Reduction for floating-point scientific data on structured and unstructured grids. With exceptional data compression capability and precise error control, MGARD addresses a wide range of requirements, including storage reduction, high-performance I/O, and in-situ data analysis. It features a unified application programming interface (API) that seamlessly operates across diverse computing architectures. MGARD has been optimized with highly-tuned GPU kernels and efficient memory and device management mechanisms, ensuring scalable and rapid operations.
\end{abstract}

\begin{keyword}
Error-controlled data compression \sep Data refactoring \sep I/O acceleration \sep Derived quantities preservation



\end{keyword}

\end{frontmatter}


\noindent
\section{Motivation and significance}
In today's scientific landscape, large-scale scientific applications generate an overwhelming volume of data, surpassing the capabilities of network and storage systems. For instance, 
the Square Kilometer Array (SKA) telescope, designed to explore radio-waves from the early universe, is projected to deliver around 600 Petabytes of data per year to a network of SKA Regional Centers for ingestion and storage \citep{sanchez2021ska}. 
Despite this data deluge, modern parallel file systems (PFS) exhibit limited aggregated bandwidth, typically measured in several Terabytes per second. The throughput of Wide Area Network (WAN) for long-distance data transmission is even sluggish, usually in the range of several hundred Megabytes per second. 
A parallel trend has also emerged in artificial intelligence community, marked by growing demands for storage and memory resource to support the training of increasingly deeper, wider, and non-linear deep neural networks (DNN). Additionally, the efficiency of DNN operations is hindered by rising communication costs associated with sharing model parameters during distributed training.

Compression has emerged as a promising solution to address the challenges posed by storage and I/O bandwidth limitations. The ideal compression approaches seek to reduce data size by several orders of magnitude while preserving its fidelity for reliable scientific use. 
The ability to refactor data into a multi-scale representation that aligns with the hierarchical architecture of storage tiers is also highly desirable. However, the presence of random mantissa with the floating-point representation of scientific data limits the compression ratios~\citep{son2014data, lindstrom2006fast} with conventional entropy-based lossless compressors~\citep{burtscher2008fpc, collet2021rfc, deutsch1996gzip, NVCOMP}. Alternative approaches, like sparse output rate compression, have their limitations too, potentially overlooking valuable scientific insights in unsaved timesteps.

Recently, lossy compression has garnered increased attention due to its effectiveness in reducing data stored in floating-point precision. A typical lossy compressor involves decorrelation, precision truncation, and lossless encoding steps, along with mathematical theories to control data distortion. An ideal lossy compressor for scientific data reduction should possess the following features: (1) strict error control with respect to different norms, (2) high throughput to avoid I/O bottlenecks, (3) portability on mainstream computing processors, (4) the ability to handle data defined on various grid structures, and (5) the capability to refactor data into multi-scales.

In this regard, several state-of-the-art lossy compressors have been developed. SZ~\citep{zhao2021optimizing}, ZFP~\citep{lindstrom2014fixed}, TTHERSH~\citep{ballester2019tthresh}, and FPZIP~\citep{lindstrom2017fpzip} offer APIs accepting $L^2$ or/and $L^\infty$ error bound settings. 
SZ offers additional error controls for several types of quantities of interest (QoIs), including polynomials, logarithmic mapping, weighted sum, and critical point/isosurface \citep{liang2022toward,jiao2022toward}. In terms of the throughput, although 
SZ and ZFP provide high-performance libraries--cuSZ~\citep{tian2020cusz} and cuZFP \citep{cuZFP}--on NVIDIA GPUs, they only support single precision and fixed-rate compression mode separately, resulting in limited usability and lower compression ratios. Moreover, these GPU-based compressors lack out-of-core support, requiring users to manually tile and fit data into GPU memory, impacting throughput performance. Additionally, existing error-bounded lossy compressors (e.g., SZ, ZFP, FPZIP, TTHERSH) are limited to data defined on uniformly spaced grids up to four dimensions.

Addressing these challenges, our present paper describes MGARD: the MultiGrid Adaptive Reduction for Data~\citep{ainsworth2018multilevel,ainsworth2019multilevel,ainsworth2019qoi} a high-performance framework designed for compressing and refactoring scientific data defined on various grid structures while ensuring precise error control. By decomposing floating-point data into a hierarchical representation on multigrid and applying quantization, MGARD achieves exceptional compression capabilities for scientific data. Importantly, the induced information loss during compression is mathematically guaranteed by finite element theories, ensuring the trustworthiness of the compressed data for a wide range of scientific applications. 
MGARD offers refactoring functionality as an alternative to lossy compression for applications requiring near-lossless storage and the flexibility to access data in various scales. It supports refactoring data into a set of components representing hierarchical resolutions and precision, enabling users to incrementally retrieve and recompose them to any accuracy on demand. Moreover, MGARD's state-of-the-art implementation supports compressing and refactoring data defined on various mesh topologies and offers multi-resolution and multi-precision parametrization options. It delivers high performance and scalability on leadership high-performance computing (HPC) facilities, such as Summit and Frontier. Previous works have shown that the high-throughput compression on GPU helps accelerate the training of large-scale DNNs by reducing the communication latency~\citep{zhou2023accelerating}. Furthermore, DNNs trained using data reduced by error-bounded compressors exhibit little or no accuracy loss~\citep{grabek2021impact,jin2021comet}.

MGARD consists of GPU and CPU kernels. Implemented in C++11~\citep{stroustrup2013c++}, OpenMP~\citep{openmp}, CUDA~\citep{cuda}, HIP~\citep{hip}, and SYCL~\citep{sycl}, MGARD leverages platform portability and embraces modern software engineering practices, including unit testing and continuous integration. The framework provides a unified application programming interface (API) with a level of abstraction focused on data reduction and reconstruction in scientific workflows. With built-in compile-time auto-tuning and runtime adaptive scheduling techniques, users can expect the best performance across different computing architectures. 
MGARD is part of the United States Department of Energy (DOE) Exascale Computing Project (ECP) software technology stack for data reduction~\citep{kothe2018exascale,messina2017exascale}, which solidifies its position as a crucial component in the advancement of data reduction technologies.

\section{Software design}
\begin{figure}[H]
    \centering
    \includegraphics[width=\textwidth]{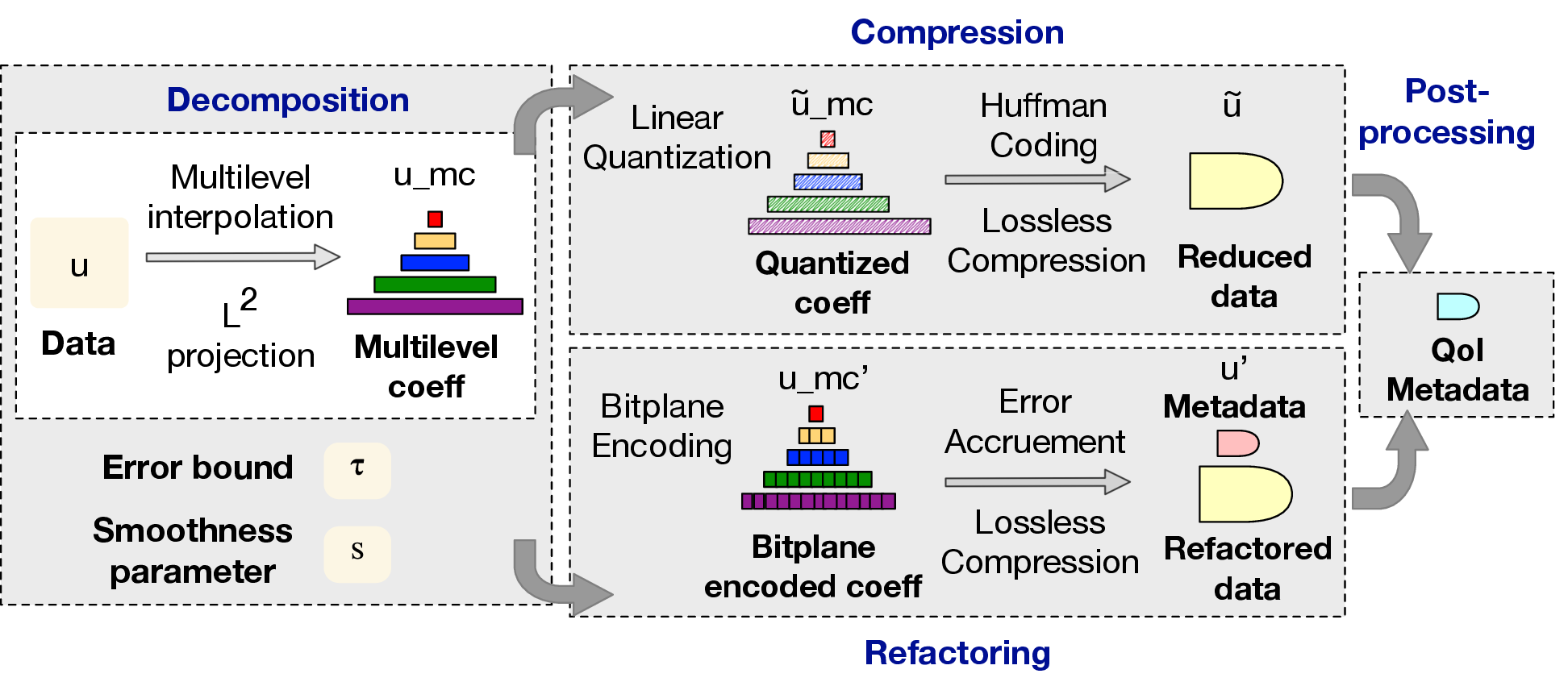}
    \caption{The software pipeline overview illustrating the two primary functionalities of MGARD -- compression and refactoring, both with precision error control. }
    \label{fig:MGARD_softwareFamily}
\end{figure}

As illustrated in Figure~\ref{fig:MGARD_softwareFamily}, the inputs to MGARD API consist of a data array \texttt{u}, user-prescribed error bound(s) $\tau$, and a smoothness parameter \texttt{s}, which defines the norm of error metrics. 
MGARD comprises two primary modules: data compression and refactoring. Both modules start with a common practice, recursively decomposing \texttt{u} into a sequence of approximations at various levels of the multi-resolution hierarchy. This decomposition generates a multilevel representation, $\uMultilevelCoefficients$, which is better suited for compression and refactoring processes.

The compression module involves a quantization stage where each component of $\uMultilevelCoefficients$ is approximated by a multiple of a quantization bin width~\citep{tao2017significantly, liang2021mgard+}. This linear quantization effectively transforms floating-point data into integers, facilitating efficient coding and ensuring that the specified error bound for $\uMultilevelCoefficients$ is met. On the other hand, the refactoring module encodes $\uMultilevelCoefficients$ into precision segments with varying significance at different levels of the multi-resolution hierarchy, utilizing bitplane encoding \citep{schwartz1966bit}. Both compression and refactoring modules employ the same set of error estimators for accuracy control, which are analogous to the posterior error estimators used in numerical analysis. These error estimators consider quantization errors or precision segments of multilevel coefficients as inputs, allowing error control in various metrics and linear Quantities of Interest (QoIs) \citep{ainsworth2019multilevel, ainsworth2018multilevel, ainsworth2019qoi}.

In the final stage, the quantization and precision segments obtained from the compression and refactoring modules are compressed through lossless encoding and written to disk as a self-describing buffer containing all the necessary parameters for decompression and recomposition. The compressed/refactored representation may also undergo post-processing, especially for preserving non-linear QoIs. The refactoring module includes an additional step that accumulates errors in the precision segments of the multilevel coefficients. The recomposition module, operating in an inverse procedure to refactoring, employs a greedy algorithm to determine the retrieval order of precision segments. This strategy aims to fetch the most significant segment across all levels based on the previously accrued error estimators.

\subsection{APIs}
MGARD is designed with two levels of APIs to support the integration with different user applications and IO libraries.

\subsubsection{High-Level APIs}
The high-level APIs offer an all-in-one compression and refactoring solution, providing users with a seamless integration experience with MGARD. Key features of the high-level APIs include:

\begin{itemize}
    \item Unified APIs: MGARD offers a single set of APIs for compressing and refactoring. MGARD automatically optimizes the reduction and refactoring kernels for the targeted GPU or CPU architectures during the software installation stage. Users only need to integrate with MGARD once to utilize it across various systems, enhancing code portability and ease of use.
    \item Self-describing format: The output of compression and refactoring APIs includes all the necessary information required by a decompressor/re-compositor to read and reconstruct data correctly. This encompasses vital details such as the compressor’s version, error bounds employed, data topologies, and the type of lossless encoders utilized. 
    \item Unified memory buffers on CPU/GPUs: MGARD automatically detects the locations of input/output buffers and handles the host-to-device data transfer internally, eliminating the need of manual setup.
    \item Multi-device out-of-core processing: The high-level APIs can automatically detect and leverage multiple accelerator devices on a system. MGARD also boasts with an out-of-core optimization to manages memory overflow and inter-device data transfer. These functionalities are crucial for large-scale data processing, where GPUs often have smaller memory capacities compared to CPU hosts. 

\end{itemize}
\subsubsection{Low-Level APIs}
The low-level APIs offer users complete control over the compression and refactoring processes, empowering them to customize the functionality based on their specific application needs. Key features of the low-level APIs include: 
\begin{itemize}
    \item Highly customizable code pipeline: The low-level APIs expose individual functions for each step within compression and refactoring, such as memory management and sub-operations. This level of granularity allows users to construct their own highly optimized compression/refactoring pipelines tailored to their application's requirements.
    \item Device asynchrony: The low-level APIs allow users to pipeline computation and cross-device data transfer so they will execute asynchronously. For example, overlapping MGARD operations on GPUs with the application's workload on CPUs. This opens up significant opportunities for users to optimize MGARD in tandem with their application's execution logic, leading to enhanced performance.
\end{itemize}

The dual-tiered API approach of MGARD ensures that users seeking a quick and easy integration with minimal effort and those requiring granular control over the compression and refactoring processes are both catered. 

\subsection{Software architecture}
\begin{figure}[H]
    \centering
    \includegraphics[width=1\textwidth]{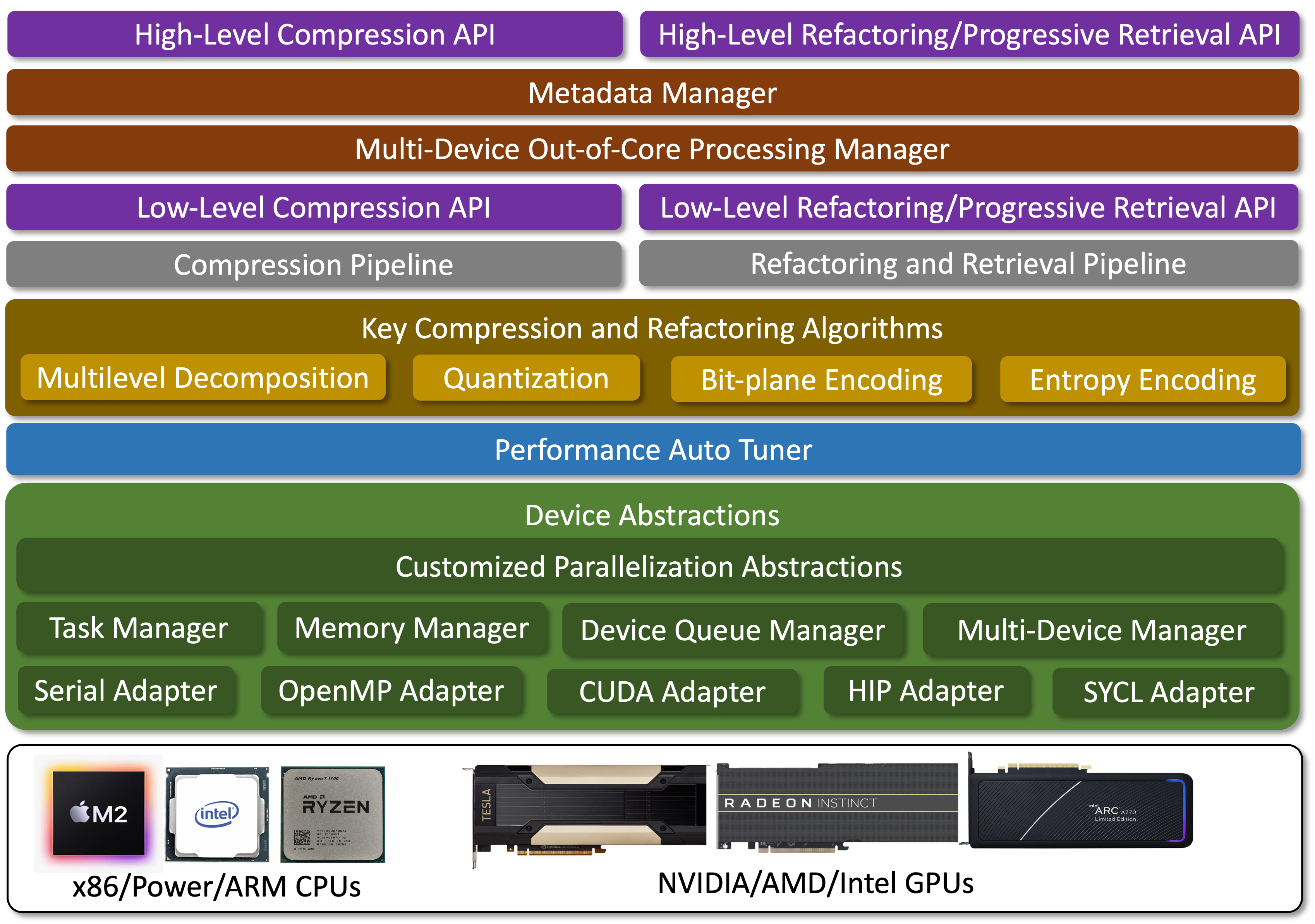}
    \caption{Software architecture of MGARD}
    \label{fig:Software_Architecture}
\end{figure}

MGARD is meticulously designed to be highly functional, performant, portable, and extendable, achieved through a modularized software architecture with carefully designed abstraction layers for maximum portability. It has been successfully integrated into ADIOS--a high-performance parallel I/O framework with an extensive user community--as an inline compressor. This integration allows ADIOS users to write and compress data using MGARD in a single step. Figure~\ref{fig:Software_Architecture} provides an illustration of MGARD's software architecture. 

At the foundation of the architecture are device abstractions (green), which ensure the sustained functionality irrespective of underlying hardware features. 
One layer above (blue), MGARD incorporates a built-in auto-tuning module. This model automatically adjusts performance configurations, such as GPU thread block sizes, shared memory usage, and processor occupancy in the software installation stage, ensuring that MGARD operates efficiently on targeted hardware micro-architectures. 
The design of MGARD's auto-tuning module draws inspiration from techniques discussed in \citep{jiang2005automatic,tillet2017input,li2009note,cuenca2004architecture,whaley1998automatically}, primarily focusing on optimization at the kernel functions level. 
The subsequent layer (dark yellow) houses the central computation kernels used by the compression and refactoring processes. They serve as the foundational building blocks for MGARD's compression and refactoring pipelines (gray) to assemble with. These functionalities are exposed to users through a set of low-level APIs. These low-level APIs offer users the ability to fine-tune the process according to their specific application needs. 
The separation between the low-level and high-level APIs is marked by the inclusion of out-of-core processing and metadata management (dark red). The out-of-core processing dynamically partitions data arrays into multiple chunks that fit within the device memory, allowing users to feed MGARD with arbitrarily large input array. On the other hand, the metadata management layer saves all information required for data reconstruction and recomposition in a self-describing format. 
The high-level APIs encapsulates underlying complexity into a single line of code for compression, decompression, refactoring, and recomposition separately (as illustrated in examples later presented in Section \ref{sec:API_examples}). It enable users to easily integrate MGARD into their applications without delving into the intricacies of the data reduction and refactoring processes. 

\subsection{Software functionalities}
\label{sec:software-fun}
MGARD primarily focuses on two functionalities: compression and refactoring, and mathematically guarantees that the information loss induced by compression and refactoring adheres to user-prescribed error tolerance. The compression functions can promote scientific discoveries by releasing storage burden so simulation/devices can output data at enhanced resolutions/frequencies~\citep{gong2023region}. They could also accelerate I/O due to MGARD's high-throughput on GPUs. 
As data volumes and velocities continue to increase, scientists require tools to incrementally retrieve, move, and process reduced data based on scientific priorities and resource constraints. MGARD's refactoring functionality empowers users to make trade-offs among uncertainty, speed, and resource utilization. 
Furthermore, scientific data often undergoes a process where it is compressed at one place/device and then transferred to different sites/devices for various analyses. MGARD's unified API facilitates cross-platform data sharing through its design, encompassing functions, and format portability.

\section{Illustrative examples}
\label{sec:API_examples}
The following examples illustrate MGARD's compression and refactoring APIs. MGARD employs the same set of APIs for backend functions running on various GPU and CPU architectures and will automatically switch to the most optimal processors available. 

Listing 1 showcases MGARD's high-level APIs for compression and reconstruction. Users provide the error bound, error metric, and the smoothness parameter as the inputs. The resulting compression ratio is obtained by dividing \texttt{in\_byte} and \texttt{out\_byte}. One noteworthy aspect is that MGARD's interface automatically detects the available device memory and location of buffers holding \texttt{in\_array}, \texttt{compressed\_array}, and \texttt{decompressed\_array}. When GPUs devices are used, the high-level APIs dynamically schedules the out-of-core processing and manages host-to-device data transfer internally.
\begin{lstlisting}[language=C, caption={MGARD data compression and decompression API example}]
#include "mgard/compress_x.hpp"

// prepare data buffers
mgard_x::DIM num_dims = 3;
mgard_x::SIZE n1, n2, n3;
std::vector<mgard_x::SIZE> shape{n1, n2, n3};
mgard_x::SIZE in_byte = n1 * n2 * n3 * sizeof(double);
mgard_x::SIZE out_byte;
//... load data into in_array
double *in_array = ...; 
void *compressed_array = NULL;
void *decompressed_array = NULL;
// tol: error tolerance
// s: smoothness parameter
double tol = 0.01, s = 0;

// MGARD config parameters
mgard_x::Config config;

// Compressing with high level API
mgard_x::compress(num_dims, mgard_x::data_type::Double, shape, tol, s, mgard_x::error_bound_type::REL, in_array, compressed_array, out_byte, config, false);

// Decompressing with high level API
mgard_x::decompress(compressed_array, out_byte, decompressed_array, config, false);

\end{lstlisting}

Listing 2 demonstrates how to refactor and incrementally recompose data using MGARD's high-level APIs. 
The refactoring API generates a metadata file and multi-resolution precision segments in a compressed format. 
Lines 23-42 illustrate that the recomposition process commences with a coarse representation of the data, retrieving only the partial segments that lead to the next level of precision/resolution in subsequent rounds.

\begin{lstlisting}[language=C, caption={MGARD data refactoring API example}]
#include "mgard/mdr_x.hpp"
...

// prepare data buffers
mgard_x::DIM num_dims = 3;
mgard_x::SIZE n1, n2, n3;
std::vector<mgard_x::SIZE> shape{n1, n2, n3};
mgard_x::SIZE in_byte = n1 * n2 * n3 * sizeof(double);
mgard_x::SIZE out_byte;
//... load data into in_array
double *in_array = ...; 

mgard_x::Config config;
mgard_x::MDR::RefactoredMetadata refactored_metadata;
mgard_x::MDR::RefactoredData refactored_data;

// Refactor with high level API
mgard_x::MDR::MDRefactor(D, mgard_x::data_type::Double, shape, in_array, refactored_metadata, refactored_data, config, false);

// Save refactored_metadata and refactored_data to files
...

mgard_x::MDR::ReconstructedData reconstructed_data;
// Read in refactored_metadata from file 
...
// Progressively reconstruct for each error bound
for (double tol : tolerances) {
    // Specify error bound and smoothness parameter for each subdomain
    for (auto &metadata : refactored_metadata.metadata) {
          metadata.requested_tol = tol;
          metadata.requested_s = s;
    }
    // Determine required data compoenents for reconstruction
    mgard_x::MDR::MDRequest(refactored_metadata, config);
    // Read in required data compoenents from files 
    ...
    // Reconstruct with high level API
    mgard_x::MDR::MDReconstruct(refactored_metadata, refactored_data, reconstructed_data, config, false, original_data);
    
    // reconstructed_data now contains progressively reconstructed data
    double out_data = reconstructed_data.data;
}
\end{lstlisting}


\section{Application impact}
The MGARD team has worked with application scientists from a variety of research communities to demonstrate MGARD's functionalities. 
\subsection{Plasma physics}
\begin{figure}[H]
    \centering
    \includegraphics[width=0.55\textwidth]{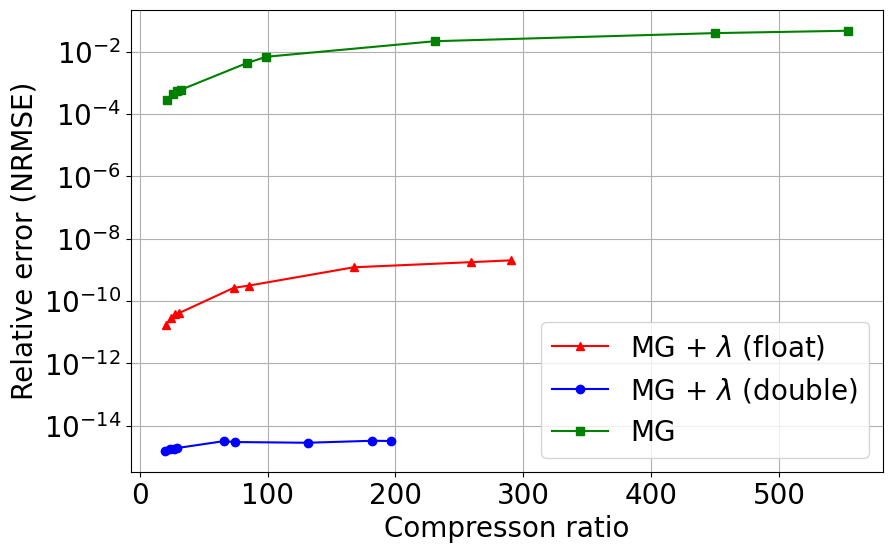}
    \caption{Illustration of errors in QoIs derived from the XGC f-data lossy compressed by MGARD with QoI post-processing.}
    \label{fig:MGARD_Lambda_XGC}
\end{figure}

\begin{itemize}
    \item [--] \textbf{XGC:} The X-point included Gyrokinetic Code (XGC) is a fusion physics code specialized in simulating plasma dynamics in the edge region of a tokamak reactor \citep{chang2008spontaneous,ku2009full}. We compressed the 5D particle distribution function (pdf) generated by XGC simulating an ITER-scale experiment~\citep{claessens2020iter}, and evaluated the errors in five derived QoIs (density, parallel/vertical temperatures, and two flux surface averaged momentums). Figure \ref{fig:MGARD_Lambda_XGC} illustrates that the MGARD with QoI post-processing can reduce the data storage for up to $200\times$ and $290\times$ with the relative $L^2$ errors in all QoIs below \num{1e-14} and $\num{1e-8}$ separately, whereas the compression without QoI optimization exhibits a relative $L^2$ error of approximately \num{1e-2} given the same compression ratios. Noted that $\lambda$ represents the set of Lagrange multipliers obtained through a convex optimization program aiming to reduce QoI errors in each sub data-domain. $\lambda$ can be further quantized or truncated to increase compression ratios. 
    Readers can find more MGARD studies on XGC simulation data in \citep{gong2021maintaining,lee2022error,banerjee2022algorithmic}. 
\end{itemize}

\begin{figure}[H]
    \centering
    \includegraphics[width=\textwidth]{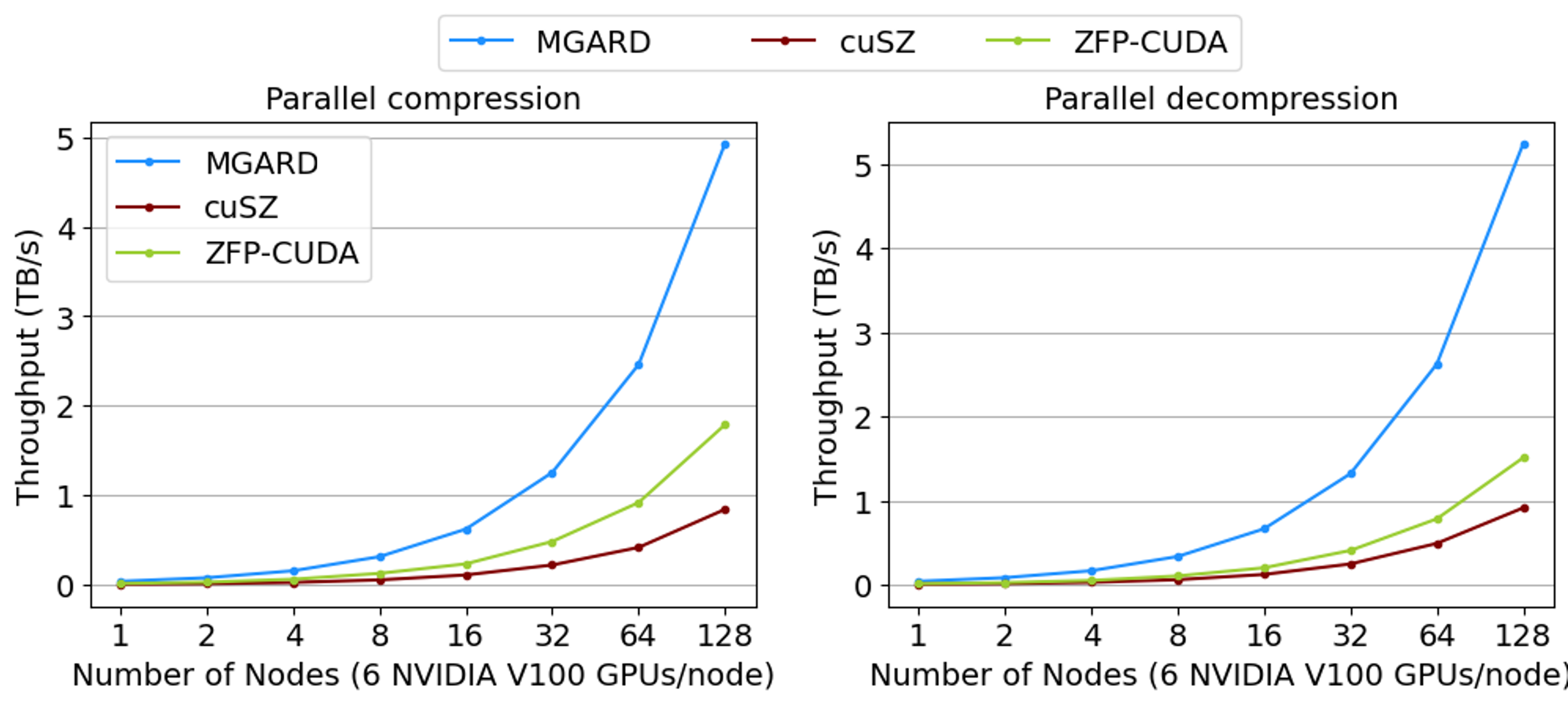}
    \caption{Comparing the throughput performance of compression and decompression provided by MGARD, cuSZ, and ZFP-CUDA on OLCF Summit nodes, using NYX data and a relative error bound of \num{1e-3}.}
    \label{fig:GPU_speed_NYX}
\end{figure}

\subsection{Earth and cosmological science}
\begin{figure}[h]
    \centering
    \begin{subfigure}[b]{0.75\linewidth}
    \centering
    \includegraphics[width=\textwidth]{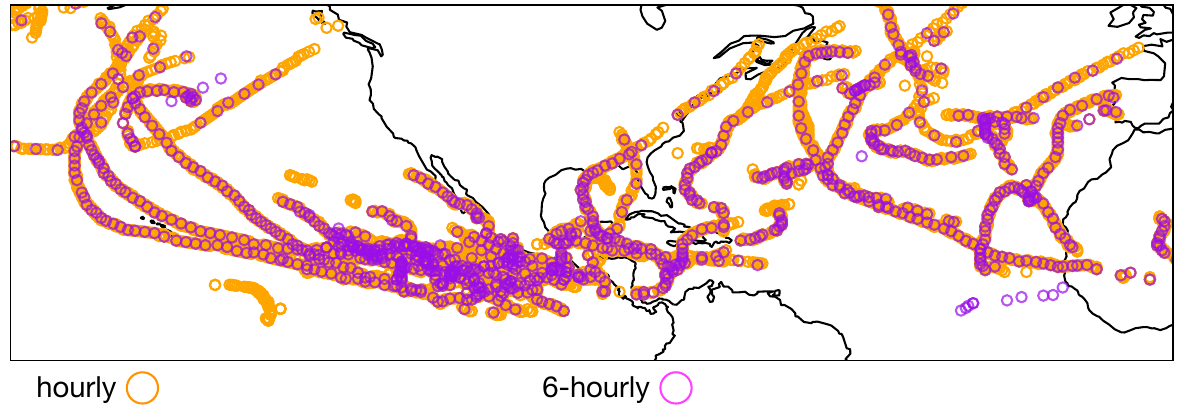}
    \caption{Visualize the TC tracks found in hourly and 6-hourly data (temporal decimation rate = 6).}
    \label{fig:TC_uniform}
    \end{subfigure}
    \hfill
    \begin{subfigure}[b]{0.75\linewidth}
    \centering
    \includegraphics[width=\linewidth]{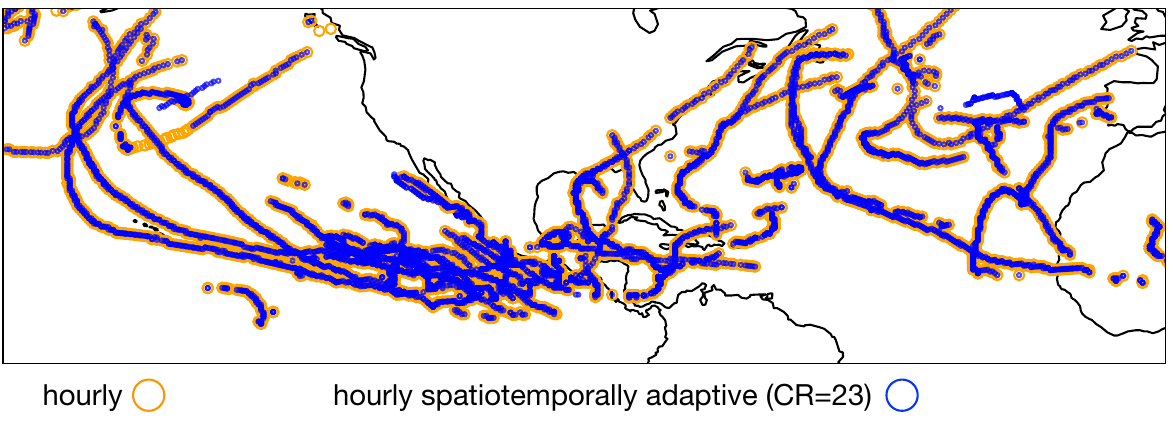}
    \caption{Visualize the TC tracks found in hourly and hourly spatiotemporally compressed data (compression ratio = 23).}
    \label{fig:TC_adaptive}
    \end{subfigure}
    \caption{Global distributing of TC tracks detected in hourly, 6-hourly, and spatiotemporally adaptive reduced hourly data over one year time span.}
    \label{fig:TC_qualitative_eval}
    \end{figure}

\begin{itemize}
    \item [--] \textbf{NYX:} NYX is an AMR-based cosmological hydrodynamics simulation code developed at Lawrence Berkeley National Laboratory \citep{sexton2021nyx}. Figure \ref{fig:GPU_speed_NYX} presents the compression and decompression throughput of MGARD and the GPU implementation of two other state-of-the-art lossy compressors: cuSZ and ZFP-CUDA. The throughput data was obtained from the Summit supercomputer~\citep{summit}, where each compute node hosts six NVIDIA V100 GPUs. For our evaluation, we fed each GPU with 15GB of NYX data, using a relative $L^2$ error bound of \num{1e-3} for data compression. Throughout the evaluation, MGARD surpassed other GPU-accelerated lossy compressors in terms of performance due to its efficient compression kernels and multi-GPU pipeline optimization. Figure \ref{fig:GPU_IO_NYX} illustrates how data compression accelerated I/O throughput in NYX simulations. Using the same setting as the experiments in Figure~\ref{fig:GPU_speed_NYX}, we compare the combined 
    time spent on compression/decompression and reading/writing the reduced data against the time spent on reading and writing the uncompressed data. The results suggest that data compression can effectively reduce the I/O cost, and MGARD exhibits the most significant improvement 
    among the three lossy compressors.  

\begin{figure}[H]
    \centering
    \includegraphics[width=\textwidth]{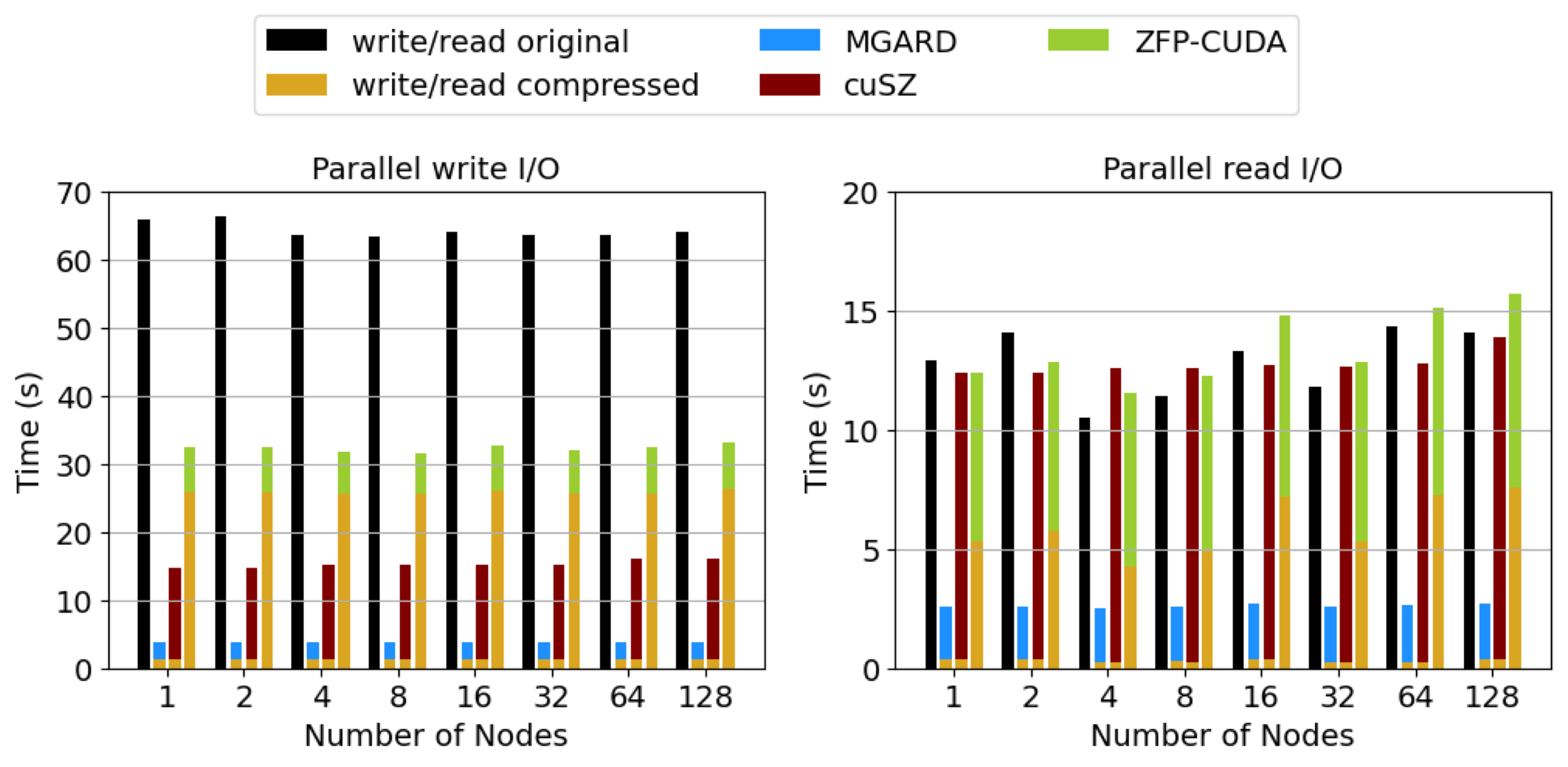}
    \caption{Comparing the end-to-end I/O time for reading and writing both compressed and uncompressed NYX data using MGARD, cuSZ, and ZFP-CUDA. Each node accommodates six NVIDIA V100 GPUs.}
    \label{fig:GPU_IO_NYX}
\end{figure}

\item [--] \textbf{E3SM:} The Energy Exascale Earth System Model is a state-of-the-science Earth's climate model used to investigate energy-relevant science~\citep{caldwell2019doe}. Due to storage constraints, E3SM currently outputs model data at 6-hourly intervals instead of the physical timestep, which is 15 minutes. In Figure \ref{fig:TC_adaptive} \citep{gong2023region}, the tropical cyclone (TC) tracks detected from data outputted at hourly intervals are compared with TC tracks obtained from the same set of data, lossy compressed using MGARD with distinct error bounds tailored to regions with varying degrees of turbulence. Concurrently, Figure~\ref{fig:TC_uniform} illustrates TC tracks detected from data outputted at a 6-hourly rate. Remarkably, despite the lossy compression of hourly data requiring only $1/4$ of the storage compared to the uncompressed 6-hourly data, a notable enhanced accuracy is achieved. 
\end{itemize}
    
\subsection{Radio astronomy}
\begin{itemize}
    \item [--] \textbf{SKA:} The Square Kilometre Array (SKA) \citep{van2015casacore} hosts two of the world's largest radio telescope arrays, archiving approximately 300 petabytes of data per year. Early exploration work has indicated that MGARD can compress radio astronomy data by approximately $20
    \times$ without introducing structural artifacts. Ongoing efforts aim to integrate data reduction into the Casacore Table Data System's I/O pipeline. 
\end{itemize}

By showcasing the impact of MGARD in diverse applications, it is evident that MGARD significantly tackles data storage and I/O challenges in the workflow of large-scale scientific experiments while ensuring the preservation of vital scientific insights. 

\section{Conclusion}

MGARD has been designed to tackle storage, I/O, and data analysis challenges for scientific applications. With novel multilevel decomposition, advanced encoding, and rigorous error control techniques, MGARD can compress data into a greatly reduced representation or refactor the data into a format supporting incremental retrieval. A well-developed mathematical foundation allows MGARD to provide error bounds not just on the raw data but also on QoIs derived from the lossy reduced data. With the mathematically proved theories and solid empirical evaluations, MGARD provides compression that will not compromise the scientific validity and utility of data. The refactoring capability of MGARD serves as an alternative to the single-error-bounded compression for users who require near-lossless data storage but may retrieve data at varied precisions/resolutions. Beyond trustworthiness, MGARD can accelerate data movement and in-situ data analytics with its extensively optimized CPU and GPU implementations, and is portable so that data compression and refactoring can operate on mainstream computing processors.

\section*{Acknowledgements}
This research was supported by the Exascale Computing Project CODAR, SIRIUS-2 ASCR research project, the Laboratory Directed Research and Development Program of Oak Ridge National Laboratory (ORNL), and  the Scientific Discovery through Advanced Computing (SciDAC) program, specifically the RAPIDS-2 SciDAC institute. 
This research used resources of the Oak Ridge Leadership Computing Facility, which is a DOE Office of Science User Facility.





\bibliographystyle{elsarticle-num} 
\bibliography{elsarticle}






\section*{Current code version}
\label{}
\begin{table}[!h]
\begin{tabular}{|l|p{6.5cm}|p{6.5cm}|}
\hline
\textbf{Nr.} & \textbf{Code metadata description} & \textbf{Please fill in this column} \\
\hline
C1 & Current code version & 1.5.1 \\
\hline
C2 & Permanent link to code/repository used for this code version & github.com/CODARcode/MGARD \\
\hline
C3  & Permanent link to Reproducible Capsule & codeocean.com/capsule/4683587 \\
\hline
C4 & Legal Code License   & Apache-2.0 license \\
\hline
C5 & Code versioning system used & git \\
\hline
C6 & Software code languages, tools, and services used & C++, CUDA, HIP, SYCL, OPENMP \\
\hline
C7 & Compilation requirements, operating environments & Software: NVCOMP, ZSTD. Hardware: NVIDIA GPU, AMD GPU, x86 CPU, ARM CPU, Power CPU\\
\hline
C8 & If available Link to developer documentation/manual & github.com/CODARcode/MGARD
/blob/master/README.md \\
\hline
C9 & Support email for questions & jchen3@uab.edu or gongq@ornl.gov\\
\hline
\end{tabular}
\caption{Code metadata (mandatory)}
\label{} 
\end{table}

\end{document}